\documentclass[runningheads]{llncs}
\usepackage[T1]{fontenc}
\usepackage{graphicx,verbatim}

\usepackage{amsmath,amsfonts,amssymb}
\usepackage{url}
\usepackage{multirow}
\usepackage{float}

\begin{document}

\title{Dual-Domain Cross-Modal Decoding for Clinical Text-Guided Medical Image Segmentation}
\titlerunning{Dual-Domain Cross-Modal Decoding}

\author{Md Maklachur Rahman \and
Tracy Hammond}
\authorrunning{M. M. Rahman and T. Hammond}
\institute{Texas A\&M University, College Station, TX 77843, USA\\
\email{\{maklachur,hammond\}@tamu.edu}}

  
\maketitle              

\begin{abstract}
Clinical text can narrow down what to segment, but recent text-guided designs emphasize spatial alignment while overlooking frequency content that governs texture and boundaries. We propose Dual-Domain Cross-Modal Decoding (DD-CMD) for clinical text-guided pulmonary infection segmentation, integrating two complementary forms of language guidance during decoding. In the spatial domain, Text-Guided Spatial Cross-Attention (TGSA) aligns multi-scale visual tokens with text semantics and updates features through gated residual fusion. In the frequency domain, Spectral-Text Adaptive Modulation (STAM) applies a 2D DCT to compute learnable band-energy statistics and predicts text-conditioned FiLM parameters to recalibrate decoder channels for frequency-aware decoding. DD-CMD embeds TGSA and STAM into a coarse-to-fine decoder ($7{\times}7 \rightarrow 56{\times}56$) and restores full-resolution masks using a lightweight two-stage refinement module. Experiments on QaTa-COV19 and MosMedData+ show that DD-CMD achieves 91.46\% Dice / 84.26\% mIoU and 81.95\% Dice / 69.42\% mIoU, respectively, with average gains of +1.96 Dice and +2.67 mIoU over the strongest prior baselines. Code: https://github.com/maklachur/DD-CMD.


\end{abstract}

\section{Introduction}
Medical image segmentation delineates anatomical structures and pathological regions and is a key step in clinical image analysis. Accurate masks support diagnosis, treatment planning, and disease monitoring across diverse applications. Encoder--decoder networks, such as U-Net~\cite{unet} and U-Net++~\cite{unet++}, are widely used baselines, while nnUNet~\cite{nnunet} improves reliability through a self-configuring training pipeline. More recently, transformer-based designs such as Swin-UNet~\cite{swinunet} strengthen long-range modeling. Despite this progress with image-only approaches, pulmonary infection segmentation remains challenging, where abnormalities can be diffuse and low-contrast, and boundaries often depend on subtle texture variations and acquisition-dependent noise.

To address these limitations, recent work has explored clinical text-guided medical image segmentation, where text provides complementary semantics about what to delineate. LViT~\cite{lvit} demonstrates that pairing images with textual descriptions and enabling cross-modal interaction can improve over vision-only baselines. Subsequent approaches refine how language interacts with visual features: MAdapter~\cite{madapter} uses adaptor-style modules to facilitate bidirectional interaction, RecLMIS~\cite{reclmis} enforces stronger cross-modal consistency via reconstruction-based conditioning, MG-UNet~\cite{mgunet} introduces a learnable memory mechanism to retain multimodal information, and ViTexNet~\cite{vitexnet} adopts text-guided dynamic operations for lightweight conditioning~\cite{mambaliteunet}. 

While these methods highlight the benefits of language, gaps remain for accurate lesion delineation. Most fusion mechanisms emphasize spatial alignment, even though texture and boundary fidelity are tightly coupled to frequency content. In addition, full-resolution cross-modal interaction can be computationally demanding, motivating coarse-to-fine designs that reserve expensive interaction for lower resolutions.

Motivated by these observations, we propose Dual-Domain Cross-Modal Decoding (DD-CMD). Our key idea is to integrate two complementary forms of language guidance during decoding: spatial alignment indicates where the model should focus, while spectral calibration determines how decoder channels respond to frequency components that govern texture and boundary structure. Concretely, DD-CMD encodes images using ConvNeXt-Tiny~\cite{liu2022convnext} and encodes clinical text using a frozen PubMedBERT~\cite{pubmedbert}. We then decode masks through three progressive stages ($7{\times}7 \rightarrow 14{\times}14 \rightarrow 28{\times}28 \rightarrow 56{\times}56$). At each stage, Text-Guided Spatial Cross-Attention (TGSA) aligns visual tokens with text semantics and updates features through a gated residual fusion. In parallel, Spectral-Text Adaptive Modulation (STAM) applies a 2D discrete cosine transform (DCT)~\cite{dct,dwtdctsvd}, aggregates learnably gated band-energy statistics, and predicts text-conditioned feature-wise linear modulation (FiLM) parameters~\cite{film} to recalibrate decoder channels in a frequency-aware manner. Finally, we recover full $224{\times}224$ resolution through a two-stage refinement module that fuses shallow spatial features while preserving text consistency.

In summary, our contributions are three-fold:
(1) We propose a dual-domain cross-modal decoder that integrates two complementary forms of language guidance: TGSA for text-guided spatial cross-attention with gated residual fusion, and STAM for frequency-aware channel recalibration using DCT band-energy statistics and FiLM conditioning.
(2) We introduce a coarse-to-fine cross-modal design that establishes image--text alignment through three decoder stages ($7{\times}7 \rightarrow 56{\times}56$) and recovers full-resolution structure with a two-stage refinement module.
(3) We conduct extensive experiments on QaTa-COV19 and MosMedData+, achieving 91.46\% Dice / 84.26\% mIoU on QaTa-COV19 and 81.95\% Dice / 69.42\% mIoU on MosMedData+, outperforming the recent baselines.
\begin{figure}[t]
    \centering
    \includegraphics[width=0.85\linewidth]{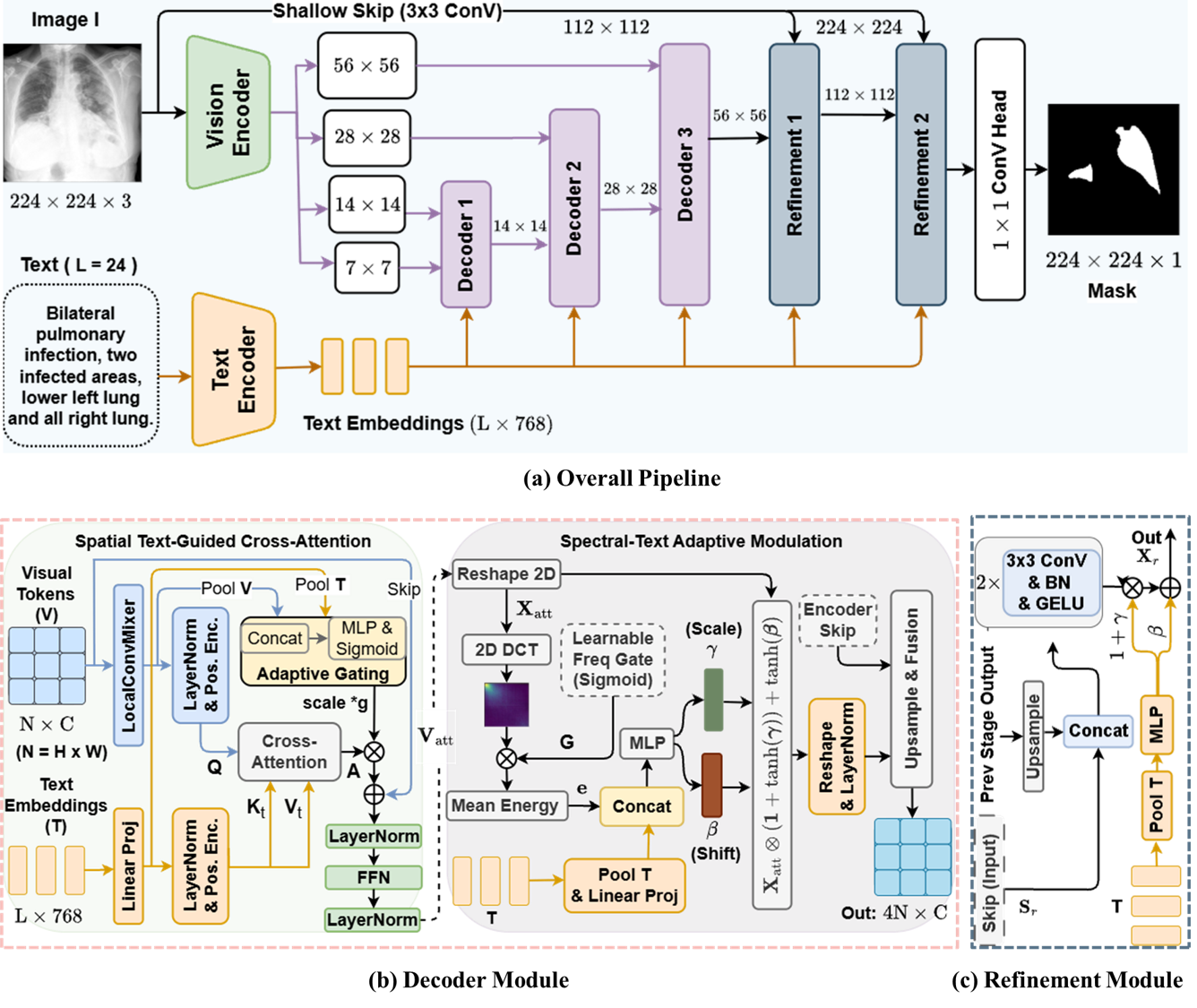}
    \caption{ (a) Overview of our proposed DD-CMD pipeline. (b) A single Decoder module. (c) A single Refinement module.}
    \label{fig:main_pipeline}
\end{figure}

\section{Methodology}
\label{sec:methodology}

An overview of DD-CMD is shown in \ref{fig:main_pipeline}. We propose DD-CMD, a text-guided medical image segmentation network that aggregates spatial cross-modal alignment with frequency-domain calibration. Given an input image $\mathbf{I}\in\mathbb{R}^{224\times224\times C}$ and a clinical text description $\mathbf{s}$, we predict a lesion mask in a coarse-to-fine decoder followed by lightweight high-resolution refinement. Our model has three key components. First, a vision encoder and a text encoder build multi-scale representations. Second, a dual domain cross-modal decoder reconstructs features from $7{\times}7$ to $56{\times}56$ for global semantic alignment. Third, a high-resolution refinement module restores full resolution from $56{\times}56$ to $224{\times}224$.

Our key design principle is to separate the two complementary roles of language guidance. Spatial attention determines where the model should focus, while spectral modulation determines how strongly each channel should respond to different frequency content. This design improves semantic alignment at low resolution and preserves fine details at high resolution.

\subsection{Vision and Text Encoders}

\subsubsection{Image encoder:}
We adopt ConvNeXt-Tiny \cite{liu2022convnext} as the vision encoder to encode the input image $I \in \mathbb{R}^{224 \times 224 \times 3}$. The encoder produces a four-scale feature pyramid
$\{F_{56},F_{28},F_{14},F_{7}\}$ at spatial resolutions
$\{56{\times}56, 28{\times}28, 14{\times}14, 7{\times}7\}$ with channel dimensions
$\{96,192,384,768\}$, respectively.
Low-resolution features capture the global context for coarse lesion localization, while high-resolution features preserve edge and texture details for better boundary reconstruction.

\subsubsection{Text encoder:}
We tokenize each clinical text, pad or truncate it to a maximum length of $L{=}24$, and encode it with a frozen PubMedBERT (base-uncased) encoder~\cite{pubmedbert}. Freezing the language branch preserves biomedical semantics and reduces overfitting when text diversity is limited. The encoder outputs token embeddings $\mathbf{T}\in\mathbb{R}^{L\times d}$ with $d{=}768$. We use $\mathbf{T}$ for cross-modal interaction and mean pooling to obtain a global descriptor $\bar{\mathbf{t}}\in\mathbb{R}^{d}$ for stage-wise conditioning.

\subsection{Dual-Domain Cross-Modal Decoder}
The decoder operates across three progressive stages: $7{\times}7 \!\rightarrow\! 14{\times}14$, $14{\times}14 \!\rightarrow\! 28{\times}28$, and $28{\times}28 \!\rightarrow\! 56{\times}56$. At each stage $s$, we process visual tokens $\mathbf{V}^{(s)}\in\mathbb{R}^{N_s\times C_s}$ (with $N_s = H_s W_s$) by two complementary components: Spatial Text-Guided Cross-Attention (TGSA) and Spectral-Text Adaptive Modulation (STAM).

\subsubsection{TGSA:}
To align visual features with text semantics while preserving local structure, we first add spatial inductive bias through a local convolutional mixer comprising depthwise $3{\times}3$ and pointwise $1{\times}1$ convolutions with BatchNorm and GELU. This step helps maintain local boundary continuity when attention flattens spatial dimensions. We then perform multi-head cross-attention where locally-mixed visual tokens serve as queries $\mathbf{Q}$ and text embeddings serve as keys $\mathbf{K_{t}}$ and values $\mathbf{V_{t}}$.
Here, $\tilde{V}^{(s)}$ denotes the output of the LocalConvMix block before cross-attention. It provides the visual queries for TGSA and is also used with text features to estimate the image-text agreement gate.

Clinical text can occasionally be vague or uninformative. To reduce over-conditioning, we introduce an adaptive gate $g^{(s)} \in (0,1)$ that scales text influence based on multi-modal agreement:
\begin{equation}
g^{(s)} = \sigma\!\left(\mathrm{MLP}\left([\mathrm{Avg}(\tilde{\mathbf{V}}^{(s)});\ \mathrm{Avg}(\mathbf{T}^{(s)})]\right)\right),
\end{equation}
\begin{equation}
\mathbf{V}^{(s)}_{\mathrm{att}} = \mathrm{LN}\!\left(\mathbf{V}^{(s)} + \alpha^{(s)} g^{(s)} \, \mathrm{MHA}(\mathbf{Q}, \mathbf{K_{t}}, \mathbf{V_{t}})\right),
\end{equation}
where $\alpha^{(s)}$ is a learnable scale parameter and $\mathrm{LN}$ denotes LayerNorm. We follow this with a feed-forward network (FFN) and another LN layer. 

\subsubsection{STAM:}
While spatial attention determines where to focus, medical images often exhibit acquisition-dependent frequency characteristics that benefit from explicit control over which frequency components to preserve. Medical anomalies exhibit distinct frequency signatures. STAM calibrates features in the frequency domain conditioned on text semantics.

Given $\mathbf{X}^{(s)}_{\mathrm{att}} \in \mathbb{R}^{C_s \times H_s \times W_s}$ (reshaped from attended tokens), we apply channel-wise 2D DCT-II~\cite{dct} to obtain frequency coefficients. Rather than fixed frequency band partitioning, we use a learnable gate $\mathbf{G}^{(s)} \in (0,1)^{H_s \times W_s}$ that softly weights informative frequency components when computing per-channel spectral energy:
\begin{equation}
e_c^{(s)} = \frac{1}{H_s W_s} \sum_{u,v} \sigma(\mathbf{G}^{(s)}_{u,v}) \left(\mathrm{DCT}(\mathbf{X}^{(s)}_{\mathrm{att}})_{c,u,v}\right)^2, \quad c=1,\dots,C_s.
\end{equation}

This forms a compact spectral descriptor $\mathbf{e}^{(s)} \in \mathbb{R}^{C_s}$. We concatenate $\mathbf{e}^{(s)}$ with the global text vector $\bar{\mathbf{t}}$ and predict FiLM parameters~\cite{film} $(\boldsymbol{\gamma}^{(s)}, \boldsymbol{\beta}^{(s)}) \in \mathbb{R}^{C_s}$ using a two-layer MLP. We apply channel-wise affine modulation and bound both scale and shift with $\tanh(\cdot)$ for stability:
\begin{equation}
[\boldsymbol{\gamma}^{(s)}, \boldsymbol{\beta}^{(s)}] = \mathrm{MLP}\!\left([\mathbf{e}^{(s)};\ \bar{\mathbf{t}}]\right),
\end{equation}
\begin{equation}
\mathbf{X}^{(s)}_{\mathrm{cal}} = \mathbf{X}^{(s)}_{\mathrm{att}} \otimes \left(1 + \tanh(\boldsymbol{\gamma}^{(s)})\right) + \tanh(\boldsymbol{\beta}^{(s)}),
\end{equation}
where $\otimes$ denotes element-wise multiplication. At the end of each decoder stage, we upsample $\mathbf{X}^{(s)}_{\mathrm{cal}}$ by $2{\times}$ via bilinear interpolation, concatenate the corresponding higher-resolution encoder skip feature $\mathbf{F}^{(s)}$, and refine the fused features using two $3{\times}3$ Conv-BN-GELU layers. Repeating this process across three stages yields text-conditioned features at $56{\times}56$.

\subsection{High-Resolution Refinement Module}
Computing full cross-attention at $112{\times}112$ and $224{\times}224$ is computationally expensive and largely redundant once global semantic alignment is established. We therefore stop cross-modal decoding at $56{\times}56$ and recover pixel-level details through two lightweight refinement blocks.

We extract shallow image features directly from the input via small convolutional layers, producing $\mathbf{S}_{112} \in \mathbb{R}^{48 \times 112 \times 112}$ (stride 2) and $\mathbf{S}_{224} \in \mathbb{R}^{24 \times 224 \times 224}$ (stride 1). These shallow pathways preserve edges and textures that are attenuated in deeper encoder stages. At each refinement resolution $r \in \{112, 224\}$, we bilinearly upsample the incoming features, concatenate the corresponding shallow skip $\mathbf{S}_r$, and apply two $3{\times}3$ Conv-BN-GELU layers. We then similarly apply lightweight FiLM conditioning from $\bar{\mathbf{t}}$ to preserve text consistency at high resolution.
Finally, a $1{\times}1$ convolution produces segmentation logits at $224{\times}224$, then a sigmoid is applied to predict the final mask.

\section{Experiments and Results}
\label{sec:experiments}
\subsubsection{Datasets, Implementation Details, and Evaluation:} We evaluate our method on two publicly available medical vision--language segmentation benchmarks, namely MosMedData+~\cite{mosmed,lvit} and QaTa-COV19~\cite{covid19,lvit}. MosMedData+ provides 2,729 COVID-19 lung CT slices with ground-truth infection masks and corresponding clinical text. QaTa-COV19 contains 9,258 chest X-ray images with lesion annotations and accompanying textual descriptions. Following~\cite{lvit,tgcam,vitexnet,fmiseg}, we adopt the exactly same train/val/test splits for fair comparison: MosMedData+ is divided into 2,183/273/273 samples, and QaTa-COV19 is split into 5,716/1,429/2,113 samples for training/validation/testing, respectively.

We implement our method in PyTorch with PyTorch-Lightning and run our experiments on an NVIDIA RTX 3090Ti GPU (24~GB). We resize all input images to $224{\times}224$ and apply random zoom, horizontal/vertical flips, and rotation, followed by intensity normalization. We train the network for 160 epochs with AdamW and cosine annealing (initial learning rate $5{\times}10^{-5}$, $\eta_{\min}=10^{-6}$) using a batch size of 8. We optimize a combined Dice and cross-entropy \cite{losssurvey} objective. Consistent with prior work~\cite{lvit,tgcam,vitexnet,fmiseg}, we report Dice and mIoU (\%) as primary metrics. We additionally assess boundary quality using HD95 (pixels) and include it in our ablations.

\subsubsection{Comparison with SOTA Methods:}
\begin{figure}[t]
    \centering
    \includegraphics[width=0.95\linewidth]{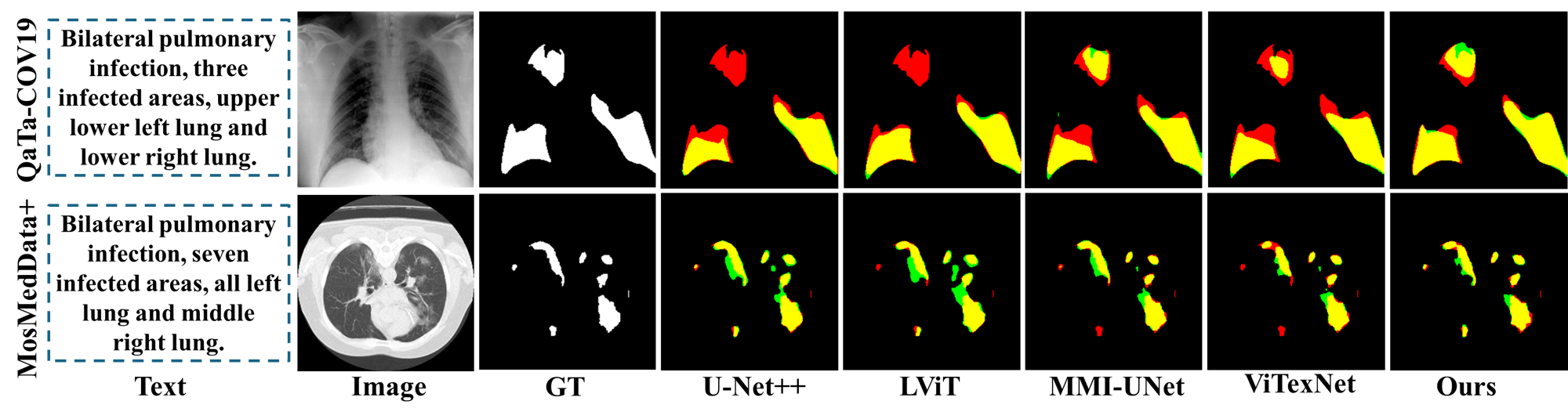}
    \caption{Qualitative comparison on QaTa-COV19 and MosMedData+. Overlays: yellow = true positives, red = false negatives, green = false positives. Best viewed zoomed in.}
    \label{fig:qc_miccai26}
\end{figure}

Table~\ref{tab:main_comp} compares text-free backbones and recent text-guided methods on QaTa-COV19 and MosMedData+. Text guidance consistently improves performance, highlighting the importance of clinical descriptions for infection delineation. DD-CMD achieves the best results on both datasets, with 91.46/84.26 Dice/mIoU on QaTa-COV19 and 81.95/69.42 on MosMedData+. Compared to MMI-UNet~\cite{mmiunet} on QaTa-COV19, DD-CMD improves by +0.58 Dice and +0.98 mIoU. On MosMedData+, it surpasses the best prior Dice by +3.33 (MAdapter~\cite{madapter}) and the best prior mIoU by +4.35 (RecLMIS~\cite{reclmis}), while using 47.77M parameters and 20.31 GFLOPs, comparable to several baselines. We also provide the qualitative comparison with the baselines in Fig.~\ref{fig:qc_miccai26}, showing better overlap on both datasets.

\begin{table*}[t]
\centering
\footnotesize
\renewcommand{\arraystretch}{1.08}
\setlength{\tabcolsep}{4.2pt}
\caption{Quantitative comparison on QaTa-COV19 and MosMedData+. Best and second-best are shown in \textbf{bold} and \underline{underlined}. Arch.: CNN, Transformer(Trans.), SAM, Hybrid (CNN--Transformer). Text indicates whether the method uses clinical texts; N/R indicates Not Reported. P and F indicate Parameters and FLOPs.}
\label{tab:main_comp}
\resizebox{\textwidth}{!}{
\begin{tabular}{l|c|c|c|cc|cc|cc}
\hline
\multirow{2}{*}{Method} & \multirow{2}{*}{Arch.} & \multirow{2}{*}{Text} & \multirow{2}{*}{Venue} &
P $\downarrow$ & F$\downarrow$ &
\multicolumn{2}{c}{QaTa-COV19} &
\multicolumn{2}{c}{MosMedData+} \\
\cline{7-8}\cline{9-10}
 &  &  &  & (M) & (G) &
Dice $\uparrow$ & mIoU $\uparrow$ &
Dice $\uparrow$ & mIoU $\uparrow$ \\
\hline
U-Net~\cite{unet}                 & CNN    & $\times$      & MICCAI'15   & 14.8  & 50.3  & 79.02 & 69.46 & 64.60 & 50.73 \\
U-Net++~\cite{unet++}             & CNN    & $\times$      & MICCAI'18   & 74.5  & 94.6  & 79.62 & 70.25 & 71.75 & 58.39 \\
nnUNet~\cite{nnunet}              & CNN    & $\times$      & Nature'21   & 19.1  & 412.7 & 80.42 & 70.81 & 72.59 & 60.36 \\
Swin-UNet~\cite{swinunet}         & Hybrid & $\times$      & ECCV'22     & 82.3  & 67.3  & 78.07 & 68.34 & 63.29 & 50.19 \\
\hline \hline
LAVT~\cite{lavt}                  & Trans. & $\checkmark$ & CVPR'22  & 118.6 & 83.8  & 79.28 & 69.89 & 73.29 & 60.41 \\
LViT~\cite{lvit}                  & Hybrid & $\checkmark$  & IEEE TMI'23 & 29.7  & 54.1  & 83.66 & 75.11 & 74.57 & 61.33 \\
LGA~\cite{lga}                    & Trans. & $\checkmark$ & MICCAI'24 & \textbf{8.24}  & 381.1 & 84.65 & 76.23 & 75.63 & 62.52 \\
MAdapter~\cite{madapter}          & CNN    & $\checkmark$  & MICCAI'24   & N/R   & N/R   & 90.22 & 82.16 & \underline{78.62} & 64.78 \\
TGCAM~\cite{tgcam}                & CNN    & $\checkmark$  & MICCAI'24   & N/R   & N/R   & 90.60 & 82.81 & 77.82 & 63.69 \\
MMI-UNet~\cite{mmiunet}           & Hybrid & $\checkmark$  & MICCAI'24   & 56.2  & 22.1  & \underline{90.88} & \underline{83.28} & 78.42 & 64.50 \\
RecLMIS~\cite{reclmis}            & Hybrid & $\checkmark$  & IEEE TMI'24 & 23.7  & 24.1  & 85.22 & 77.00 & 77.48 & \underline{65.07} \\
ARSeg~\cite{arseg}                & CNN    & $\checkmark$  & MICCAI'25   & N/R   & N/R   & 84.09 & 72.64 & 73.24 & 59.82 \\
MG-UNet~\cite{mgunet}             & Hybrid & $\checkmark$  & MICCAI'25   & 30.5  & \textbf{11.0} & 88.10 & 77.80 & 76.39 & 61.79 \\
TextMoE~\cite{textmoe}            & Hybrid & $\checkmark$  & MICCAI'25   & N/R   & N/R   & 89.08 & 80.32 & 74.66 & 59.57 \\
ViTexNet~\cite{vitexnet}          & Hybrid & $\checkmark$  & MICCAI'25   & 37.7  & 11.5  & 90.76 & 83.25 & 78.19 & 64.04 \\
\hline
\textbf{Ours} & Hybrid & $\checkmark$ & -- & 47.77 & 20.31 & \textbf{91.46} & \textbf{84.26} & \textbf{81.95} & \textbf{69.42} \\

\hline
\end{tabular}
}
\end{table*}

\subsubsection{Ablation Study on the Effect of Different Core Modules:}
Table~\ref{tab:ablation} analyzes the contribution of each component in DD-CMD. The image-only baseline performs the worst on both datasets, highlighting the benefit of clinical text for target specification. 
When we keep text but disable both TGSA and STAM, performance improves only slightly, indicating that naive conditioning is insufficient. Enabling TGSA brings a clear jump by improving text--image alignment, and adding STAM further boosts overlap while reducing boundary error, showing that spectral calibration complements spatial alignment. The high-resolution refinement primarily sharpens boundaries, while FiLM-style parameters and local mixing in TGSA bring additional consistent gains. Overall, the full model is best and most stable across datasets, showing that our decoder modules interact effectively with visual features and align lesion regions with textual semantics.

\begin{table}[t]
\centering
\footnotesize
\setlength{\tabcolsep}{3.5pt}
\caption{Effect of different core modules on QaTa-COV19 and MosMedData+. $\uparrow$/$\downarrow$ indicate higher/lower is better; $w/o$ means without. Best values are bolded.}
\label{tab:ablation}
\resizebox{\linewidth}{!}{
\begin{tabular}{l|ccc|ccc}
\hline
\multirow{2}{*}{Model Variant} &
\multicolumn{3}{c|}{QaTa-COV19} &
\multicolumn{3}{c}{MosMedData+} \\
\cline{2-7}
& Dice $\uparrow$ & mIoU $\uparrow$ & HD95 $\downarrow$
& Dice $\uparrow$ & mIoU $\uparrow$ & HD95 $\downarrow$ \\
\hline
No Text (Image-only)          & 88.17 & 78.84 & 26.90          & 79.76 & 66.33 & 18.60 \\
$w/o$ TGSA \& STAM              & 89.35 & 80.75 & 22.70          & 80.52 & 67.39 & 16.29 \\
$w/o$ TGSA                      & 90.47 & 82.60 & 16.95          & 81.01 & 68.08 & 15.03 \\
$w/o$ STAM                      & 90.92 & 83.35 & 15.71          & 81.19 & 68.34 & 14.48 \\
$w/o$ Refinement                & 91.07 & 83.60 & 14.44          & 81.35 & 68.56 & 14.12 \\
$w/o$ FiLM-Style Params ($\gamma$, $\beta$)         & 91.27 & 83.94 & 13.06          & 81.54 & 68.83 & 14.56 \\
$w/o$ Adapting Gating in TGSA                 & 91.10 & 83.65 & 14.40          & 81.41 & 68.65 & 14.09 \\
$w/o$ LocalConvMix in TGSA     & 91.33 & 84.04 & 13.44          & 81.68 & 69.03 & 14.20 \\
\hline
Full model (Ours)           & \textbf{91.46} & \textbf{84.26} & \textbf{12.76} & \textbf{81.95} & \textbf{69.42} & \textbf{13.94} \\
\hline
\end{tabular}
}
\end{table}

\begin{table}[t]
\centering
\footnotesize
\setlength{\tabcolsep}{3.5pt}
\caption{Results on varied text token length ($L$) on QaTa-COV19 and MosMedData+.}
\label{tab:ablation_textlen}

\begin{tabular}{l|ccc|ccc}
\hline
\multirow{2}{*}{text Length ($L$)} &
\multicolumn{3}{c|}{QaTa-COV19} &
\multicolumn{3}{c}{MosMedData+} \\
\cline{2-4}\cline{5-7}
& Dice $\uparrow$ & mIoU $\uparrow$ & HD95 $\downarrow$
& Dice $\uparrow$ & mIoU $\uparrow$ & HD95 $\downarrow$ \\
\hline
$L=8$   & 88.85 & 79.93 & 19.70 & 81.27 & 68.45 & 17.18 \\ 
$L=16$  & 91.22 & 83.85 & 13.28 & 81.46 & 68.72 & 14.97 \\
$L=24$ (Ours) & \textbf{91.46} & \textbf{84.26} & \textbf{12.76} & \textbf{81.95} & \textbf{69.42} & \textbf{13.94} \\
$L=32$  & 91.26 & 83.92 & 13.17 & 81.22 & 68.38 & 14.79 \\
$L=40$  & 91.17 & 83.77 & 13.80 & 80.90 & 67.93 & 16.47 \\
\hline
\end{tabular}
\end{table}
\subsubsection{Ablation Study on the Effect of Text Length (Text Tokens):}
We vary text length ($L:8\to40$) to assess how much text is needed for strong text--image alignment. Table~\ref{tab:ablation_textlen} shows that very short prompts reduce overlap and boundary quality because truncation can remove lesion descriptors needed for localization and shape. Performance improves as $L$ increases and then plateaus: $L{=}24$ provides a strong, stable default across datasets, while longer texts ($L{=}32,40$) provide no consistent gains and can slightly degrade due to diminishing returns and added noise from extra tokens. We therefore use $L{=}24$ in all main experiments for our model.

\subsubsection{Progressive Feature Visualization and Boundary Updates:}
\begin{figure}[!t]
    \centering
    \includegraphics[width=0.95\linewidth]{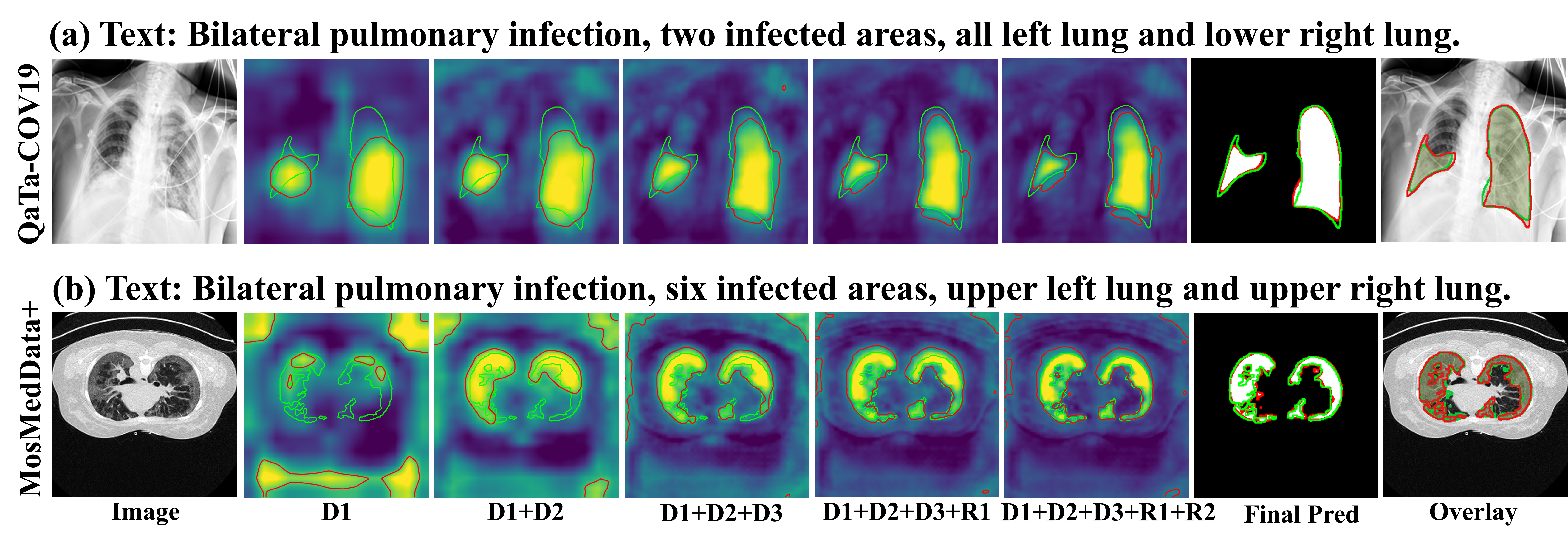}
    \caption{Progressive feature map and boundary visualization. Decoder stages (D1--D3) and refinement modules (R1--R2) are progressively added to visualize stage-wise changes. Boundaries: \textit{green = ground truth}, \textit{red = prediction}. The final columns show the predicted mask and boundary overlay. Best viewed when zoomed in.}
    \label{fig:feature_visualization_progress}
\end{figure}

Fig.~\ref{fig:feature_visualization_progress} shows how the prediction improves as we progressively add decoder and refinement blocks. Decoder (D1) gives a rough, low-resolution localization with broad activations and loose boundaries. Adding D2 and D3 strengthens lesion-specific responses and suppresses background, so the mask becomes semantically reliable and better covers the infection extent. After this point, refinement mainly improves boundaries: R1 sharpens edges and reduces leakage into nearby tissue, and R2 closes small gaps and stabilizes thin structures.

\section{Conclusion}
We proposed DD-CMD for clinical text-guided pulmonary infection segmentation, integrating spatial and spectral language guidance within a unified decoding framework. TGSA uses text-guided spatial cross-attention to align visual tokens with clinical semantics, whereas STAM calibrates decoder features through DCT band-energy statistics and text-conditioned FiLM parameters for frequency-aware decoding. This dual-domain design is implemented in a coarse-to-fine decoder ($7{\times}7 \rightarrow 56{\times}56$), with a refinement pathway to recover full-resolution masks. Experiments on QaTa-COV19 and MosMedData+ demonstrate the effectiveness of DD-CMD with strong overlap and boundary accuracy.

%
%
%
\bibliographystyle{splncs04}
\bibliography{miccai26_bibliography}

\end{document}